\documentclass[final]{cvpr}

\usepackage{times}
\usepackage{epsfig}
\usepackage{graphicx}
\usepackage{amsmath}
\usepackage{amssymb}
\usepackage{bm}
\usepackage[ruled,linesnumbered]{algorithm2e}
\usepackage{pifont}
\usepackage{booktabs}
\usepackage{array}
\usepackage{enumitem}
\usepackage{multirow}
\usepackage{xcolor}
\usepackage{amsthm}

\DeclareMathOperator{\Tr}{Tr}
\usepackage[breaklinks=true,bookmarks=false]{hyperref}
\newtheorem{proposition}{Proposition}

\def\cvprPaperID{2430} 
\def\confYear{CVPR 2021}
\begin{document}

\newcommand{\revise}[1]{\textcolor{red}{~#1}} %
\newcommand{\LY}[1]{\textcolor{red}{(Yuan:~#1)}} %
\newcommand{\WP}[1]{\textcolor{green}{(Wenping:~#1)}} %
\newcommand{\LJ}[1]{\textcolor{blue}{[\textbf{LJ}:~#1]}} 
\title{Learnable Motion Coherence for Correspondence Pruning
}

\author{Yuan Liu$^1$ \quad Lingjie Liu$^2$ \quad  Cheng Lin$^1$ \quad Zhen Dong$^3$ \quad Wenping Wang$^1$ \\[0.3em]
$^1$The University of Hong Kong \quad $^2$MPI Informatics, Saarland Informatics Campus \quad $^3$Wuhan University
}

\maketitle

\begin{abstract}
Motion coherence is an important clue for distinguishing true correspondences from false ones. Modeling motion coherence on sparse putative correspondences is challenging due to their sparsity and uneven distributions. Existing works on motion coherence are sensitive to parameter settings and have difficulty in dealing with complex motion patterns. In this paper, we introduce a network called Laplacian Motion Coherence Network (LMCNet) to learn motion coherence property for correspondence pruning. We propose a novel formulation of fitting coherent motions with a smooth function on a graph of correspondences and show that this formulation allows a closed-form solution by graph Laplacian. This closed-form solution enables us to design a differentiable layer in a learning framework to capture global motion coherence from putative correspondences. The global motion coherence is further combined with local coherence extracted by another local layer to robustly detect inlier correspondences. Experiments demonstrate that LMCNet has superior performances to the state of the art in relative camera pose estimation and correspondences pruning of dynamic scenes\footnote{Code and supplementary material can be found in the project page: \href{https://liuyuan-pal.github.io/LMCNet/}{https://liuyuan-pal.github.io/LMCNet/}}.
\end{abstract}
\section{Introduction}
Estimating correspondences between two images is a fundamental problem in computer vision tasks such as Structure-from-Motion (SfM)~\cite{hartley2003multiple}, visual localization~\cite{sattler2018benchmarking}, image stitching~\cite{brown2007automatic} and visual SLAM~\cite{mur2015orb}. 
A standard pipeline of correspondence estimation relies on the local feature matching to establish a set of putative correspondences, which contains numerous false correspondences (i.e., outliers). To prevent outliers from affecting downstream tasks, a correspondence pruning algorithm is usually applied to select a reliable subset consisting of true correspondences (i.e., inliers). 
The most prevalent correspondence pruning methods are RANSAC~\cite{fischler1981random} and its variants~\cite{torr2000mlesac,Chum2005,barath2019magsac,cavalli2020adalam}, which detect true correspondences by finding the largest subset conforming to a task-specific geometric model such as epipolar geometry or homography. However, due to large outlier ratios of putative correspondences, multiple plausible geometric models may exist, which makes it difficult for RANSAC and its variants to identify the correct geometric model. 

\begin{figure}
    \centering
    \includegraphics[width=0.5\textwidth]{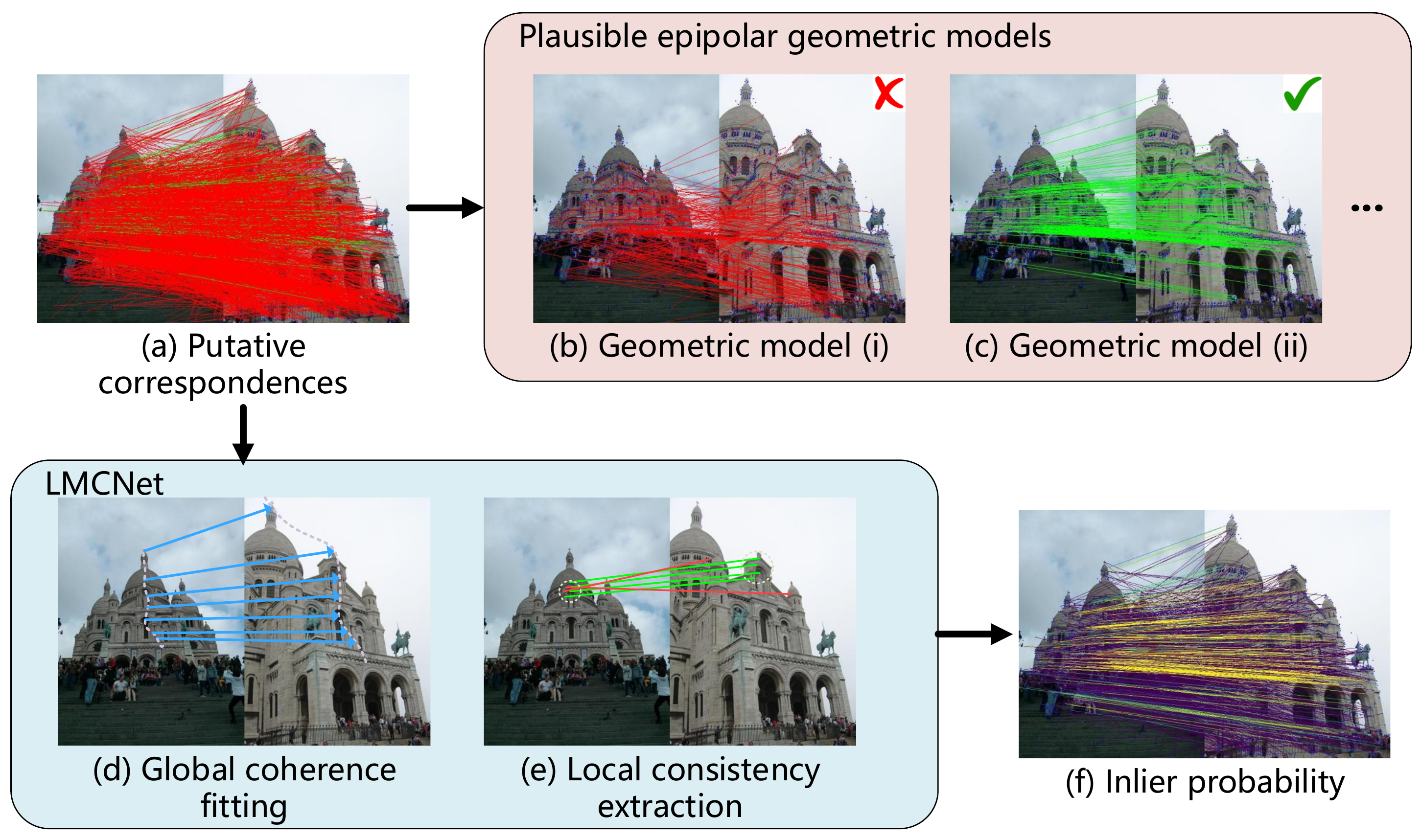} 
    \caption{Given a set of putative correspondences (a), multiple plausible epipolar geometries exist, e.g. (b) and (c). However, true correspondences (c) are usually motion coherent while false ones (b) are not. In this paper, we design a network called LMCNet to explicitly utilize the motion coherence of correspondences via a global coherence fitting (d) and a local consistency extraction (e). Hence, LMCNet is able to robustly predict their inlier probabilities (f), where brighter color means higher inlier probability. }
    \label{fig:intro}
\end{figure}
Besides the task-specific geometric model, true correspondences also conform to a more general motion model called {\em motion coherence}, which means that neighboring pixels share similar motions, while false correspondences are usually randomly scattered, as illustrated in Fig.~\ref{fig:intro} (a-c). Motion coherence supplements the task-specific geometric model and is key to determining true correspondences when multiple plausible geometric models exist.

To model motion coherence, existing works propose local~\cite{bian2017gms,ma2019locality} or global handcrafted rules~\cite{lin2014bilateral,lin2017code} to find coherent correspondences.
However, modeling motion coherence on sparse correspondences generated by local descriptors is challenging. 
First, unlike in tasks of dense correspondence estimation such as optical flow estimation~\cite{lucas1981iterative,black1993framework,black1996robust}, putative correspondences generated by local descriptors are discrete and sparse, which makes it much more difficult to estimate the underlying smooth motion field. 
Second, putative correspondences are usually distributed unevenly over an image because there are often many detected keypoints in textured regions and few keypoints in textureless regions. This uneven distribution makes it hard to find a uniform coherence constraint on the correspondences. 
Third, an observed scene may have a complex structure, such as an abrupt change of depth, so that the underlying motion fields are only piece-wise smooth, which brings about difficulty in finding motion boundaries~\cite{lin2017code}. 
Due to these challenges, existing works either need careful parameter tuning for different datasets~\cite{lin2017code,lin2014bilateral} or may fail when motion patterns are complex~\cite{ma2019locality,bian2017gms}.

We address these problems by proposing a neural network to learn the motion coherence property for correspondence pruning. Compared to handcrafted rules of motion coherence, neural networks have more powerful and flexible representational ability to learn more complex motion patterns from data and reliably detect motion boundaries. 

Designing differentiable layers to capture the motion coherence property is the key to adopting a learning-based approach. Traditional global motion coherence models~\cite{lin2017code,lin2014bilateral} usually involve a non-differentiable iterative convex optimization solver, which cannot be used for training a network end-to-end. To address this issue, we propose a novel formulation of the motion coherence property via smooth function fitting on a graph of correspondences. We call this formulation \textbf{Laplacian Motion Fitting} (LMF) and further show that the proposed LMF has a simple closed-form solution by decomposition of graph Laplacian, which enables us to design a differentiable \textbf{Coherence Residual Layer}, called CR-Layer for abbreviation, to capture global motion coherence from putative correspondences. 

Besides the global coherence model, true correspondences also have motion-consistent supporting correspondences in their local neighborhoods~\cite{bian2017gms}. 
Based on this observation, we design a \textbf{Local Coherence Layer}, called LC-Layer for abbreviation, to extract local motion coherence from these neighboring supporting correspondences. 
By integrating both the global and local motion coherence layers, we design a network called \textbf{Laplacian Motion Coherence Network} (LMCNet), which takes coordinates or other optional features of correspondences as inputs and outputs the probability of each correspondence being an inlier.

We conducted extensive experiments to demonstrate the effectiveness of the proposed neural network in two tasks: relative camera pose estimation, and
correspondence pruning of dynamic scenes. On both tasks, LMCNet achieves superior performances than other baseline methods, demonstrating its ability to robustly select inliers and potential to enhance object tracking or video object recognition.

Our contributions are as follows.
\begin{enumerate}[topsep=0pt,itemsep=-0.8ex,partopsep=0.8ex,parsep=1ex]
    \item We proposed a novel formulation of motion coherence on sparse correspondences which has a simple closed-form solution by decomposition of graph Laplacian.
    \item We proposed differentiable layers, which work together to robustly capture motion coherence of sparse putative correspondences.
    \item  We designed a neural network for correspondence pruning and demonstrate its effectiveness on the relative pose estimation problem and correspondence pruning of dynamic scenes.
\end{enumerate}

\section{Related Works}
\subsection{RANSAC-related correspondence pruning}
RANSAC~\cite{fischler1981random} and its variants~\cite{raguram2008comparative,torr2000mlesac,rousseeuw1984least,barath2018graph,raguram2012usac,barath2019magsac,barath2020magsac++,Chum2005,tiwari2018dgsac,cavalli2020adalam} are the standard methods to select true correspondences from putative correspondences by finding the largest subset which conforms to a task-specific geometric model. However, these methods may fail when multiple plausible geometric models all have a large amount of supporting correspondences. Some works address this problem by using techniques like marginalization~\cite{barath2019magsac,barath2020magsac++}, degeneration detection~\cite{Chum2005}, learning to sample hypotheses~\cite{brachmann2019ngransac} or density estimation~\cite{tiwari2018dgsac}. In this paper, we resort to another useful motion coherence property of inlier correspondences rather than solely relying on a task-specific geometric model.

\subsection{Deep networks for correspondence pruning}
With the advance of deep learning methods, pioneering works such as  DSAC~\cite{brachmann2017dsac}, PointCN~\cite{moo2018learning} and DFE~\cite{ranftl2018deep} demonstrate the feasibility of classifying correspondences by neural networks with coordinates as inputs. Follow-up works improve the architecture by inserting global clustering layer~\cite{zhang2019learning}, attention mechanism~\cite{sun2020acne,chen2019gla} or introducing new neighborhood definition~\cite{zhao2019nm}. These methods mainly focus on designing a permutation-equivariant operator on correspondences and treat the learning process as a black-box. 
In contrast, we explicitly incorporate the motion coherence property in a deep neural network to ensure that such property is learned during the training process.

\subsection{Motion coherence}
Motion coherence~\cite{yuille1989mathematical,myronenko2007non} has been explored for decades in computer vision. There are many works~\cite{lucas1981iterative,black1993framework,black1996robust,brox2010large,volz2011modeling,narayana2013coherent,rocco2020efficient,rocco2018neighbourhood,li2020dual,myronenko2007non} focusing on applying motion coherence constraints on dense correspondences estimation tasks such  as optical flow. However, fitting a smooth motion field on sparse correspondences is much harder. BF~\cite{lin2014bilateral} and CODE~\cite{lin2017code} proposed a global motion coherence model on sparse correspondences. Some other works~\cite{ma2019locality,bian2017gms} use local consistency to find motion-coherent correspondences. These handcrafted rules or models achieve impressive performances but still have difficulty in handling complex motion patterns and need careful parameter tuning for different datasets. In our method, we propose a novel formulation of motion coherence by fitting a smooth function via graph Laplacian. We further design both global and local differentiable layers to capture motion coherence, which enable our network to learn more complex motion patterns from data than those handcrafted methods.

\subsection{Image matching}
Previously, works about image matching mainly focus on learning repetitive detectors~\cite{Lowe2004Distinctive,yi2016lift,detone2018superpoint,revaud2019r2d2,barroso2019key,dusmanu2019d2,song2020sekd} or discriminative descriptors~\cite{luo2018geodesc,luo2019contextdesc,liu2019gift,tian2020hynet,tian2019sosnet,mishchuk2017working,tyszkiewicz2020disk,darmon2020learning}. However, recent works~\cite{moo2018learning} show that the performance bottleneck of image matching may be how to match these descriptors. Several deep learning-based models are proposed to learn to prune correspondences~\cite{moo2018learning,ranftl2018deep,zhang2019learning} or to match descriptors~\cite{sarlin20superglue,wiles2020d2d}. Our method belongs to the category of deep learning-based correspondences pruners.

\section{Method}
We propose a novel architecture LMCNet for correspondence pruning.
Given $N$ putative correspondences $\{\bm{c}_i=(x_i,y_i,u_i,v_i)|i=1,...,N\}$ and their optional $d_0$-dim features $\{\bm{f}_i\in \mathbb{R}^{d_0}\}$, where $(x_i,y_i)$ and $(u_i,v_i)$ are the image coordinates of two corresponding keypoints, our goal is to estimate the probability $\{p_i\}$ that a putative correspondence $\bm{c}_i$ is a true correspondence. 
In the rest of this section, we first introduce our new formulation of motion coherence, called {\em Laplacian Motion Fitting} in Sec.~\ref{sec:gcf}. Then, the key components of LMCNet, i.e. Coherence Residual Layer and Local Coherence Layer, are elaborated in Sec.~\ref{sec:crl} and Sec.~\ref{sec:lgc}. In the end, we describe the whole architecture in Sec.~\ref{sec:arc} and some implementation details in Sec.~\ref{sec:impl}.

\subsection{Laplacian Motion Fitting}
\label{sec:gcf}
Motion coherence means that true correspondences have similar motions to each other, while false correspondences scatter randomly.
Most commonly-used models~\cite{lin2017code,lin2011smoothly,yuille1989mathematical} utilize this property by first recovering the underlying continuous smooth motion field from putative correspondences. Then, true and false correspondences can be distinguished according to their deviations from the recovered motion field. However, in order to recover the underlying motion field, such formulations usually involve an iterative convex optimization solver which is non-differentiable. 
To address this, we propose a novel formulation of motion coherence by estimating a set of smooth discrete motions on a graph that encodes the adjacency of the putative correspondences.
We show that this formulation allows a simple closed-form solution via decomposition of graph Laplacian, which can be used for constructing a differentiable layer in a network.

We construct a graph $G=\{V,E\}$ where the nodes in $V$ represent all the putative correspondences, and $E$ includes the edges from every correspondence to its $k$-nearest neighbors according to their coordinate distance $d_{i,j}=\|\bm{c}_i-\bm{c}_j\|_2$. We compute the associated weights on the edges by $w_{i,j}=\exp(-d^2_{i,j}/\sigma^2)$, where $\sigma$ is a predefined constant, and $w_{i,i}=0$ for all nodes. 
Then, we define the adjacent matrix as $\bm{A}=[w_{i,j}]$, the degree matrix as $\bm{D}=diag([d_i=\sum_j w_{i,j}])$ and the Laplacian matrix as $\bm{L}=\bm{D}-\bm{A}$. 
Here, we define a matrix or a vector $\bm{v}$ by $\bm{v}=[v_{i,j}]$ whose components are $v_{i,j}$ and use $diag(\bm{v})$ to denote a diagonal matrix whose diagonal elements are the components of $\bm{v}$.

For every correspondence, we compute its motion by $\{\bm{m}_i=(m_{x,i},m_{y,i})=(u_i-x_i,v_i-y_i)\}$. 
Our goal is to estimate a set of \textit{smooth} motions $\{\bm{s}_i=(s_{x,i},s_{y,i})\}$ which are as consistent with the input motions $\{\bm{m}_i\}$ as possible. We formulate the problem as follows,
\begin{equation}
    \mathop{minimize}\limits_{\{\bm{s}_i|i=1,...,N\}}^{} \sum_i^N \|\bm{s}_i-\bm{m}_i\|^2_2+ 
    \frac{1}{2}\eta\sum_{i,j} w_{i,j}\|\bm{s}_i-\bm{s}_j\|^2_2,
    \label{eq:obj_raw}
\end{equation}
where $\|\bm{s}_i-\bm{m}_i\|^2_2$ penalizes the deviation of the estimated motion $\bm{s}_i$ from the input motion $\bm{m}_i$, $\eta$ is a predefined constant, and $w_{i,j}\|\bm{s}_i-\bm{s}_j\|_2^2$ is a smoothness cost which penalizes the motion variation between two neighboring correspondences $\bm{s}_i$ and $\bm{s}_j$ according to the weight $w_{i,j}$.

By aggregating $\bm{s}_i$ and $\bm{m}_i$ into a matrix form $\bm{s}=[\bm{s}_i]\in \mathbb{R}^{N\times 2}$ and $\bm{m}=[\bm{m}_i]\in \mathbb{R}^{N\times 2}$, we can rewrite Problem~(\ref{eq:obj_raw}) as follows,
\begin{equation}
    \mathop{minimize}\limits_{\bm{s}}^{}\  \Tr((\bm{s}-\bm{m})^\intercal(\bm{s}-\bm{m}))+\eta \Tr(\bm{s}^\intercal\bm{L}\bm{s}),
    \label{eq:obj_vec}
\end{equation}
where $\bm{s}^\intercal\bm{L}\bm{s}$ is used as a regularization term because it measures the smoothness of the graph signal $\bm{s}$~\cite{ortega2018graph}.
Problem~(\ref{eq:obj_vec}) has a closed-form solution as stated in the following proposition; we leave the proof and its connection to the previous motion coherence theories~\cite{yuille1989mathematical,myronenko2007non,lin2017code} in the supplementary material.

\begin{proposition}
Let the eigenvalue decomposition of the Laplacian matrix $\bm{L}$ be $\bm{L}=\bm{U} \bm{\Lambda} \bm{U}^\intercal$, where $\bm{\Lambda}=diag([\lambda_i])$ is a diagonal matrix of eigenvalues $\lambda_i$ and columns of $\bm{U}$ are the associated eigenvectors. Then the solution to Problem~(\ref{eq:obj_vec}) is $\bm{s}=\bm{U} diag([1/(1+\eta\lambda_i)]) \bm{U}^\intercal \bm{m}$.
\label{prop:solution}
\end{proposition}

\begin{figure*}
    \centering
    \begin{tabular}{ccc}
    \includegraphics[height=0.21\textwidth]{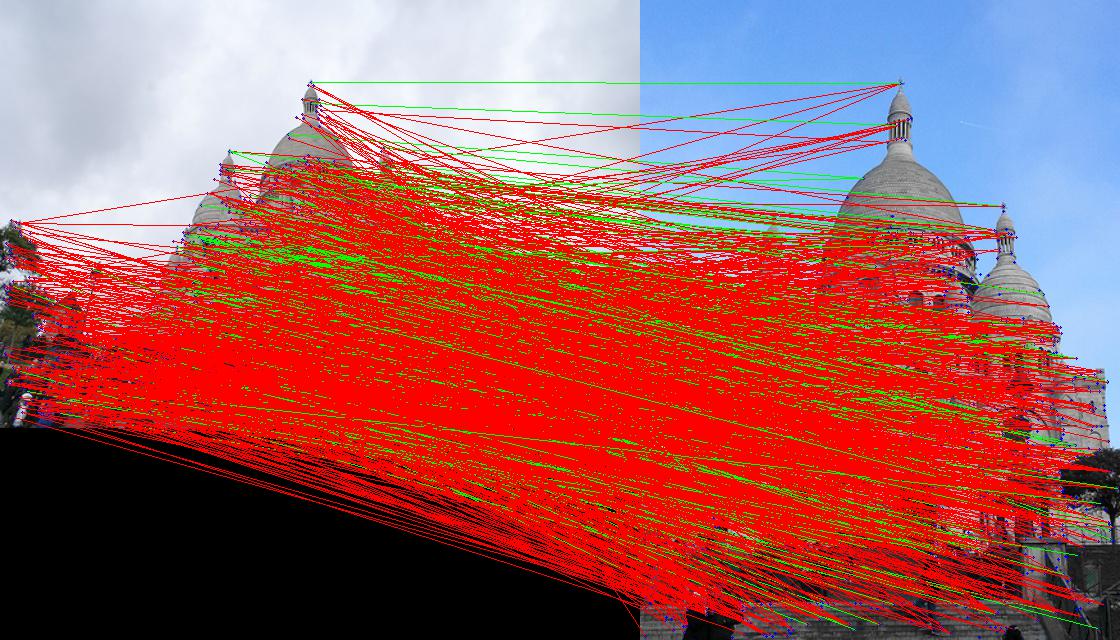} &
    \includegraphics[height=0.21\textwidth]{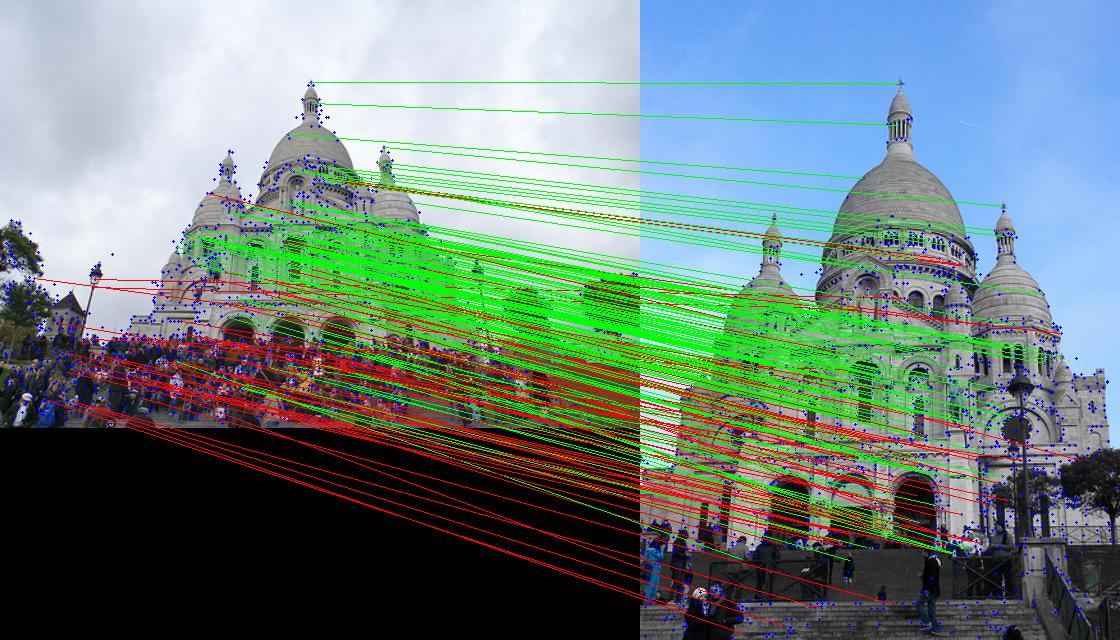} &
    \includegraphics[height=0.21\textwidth]{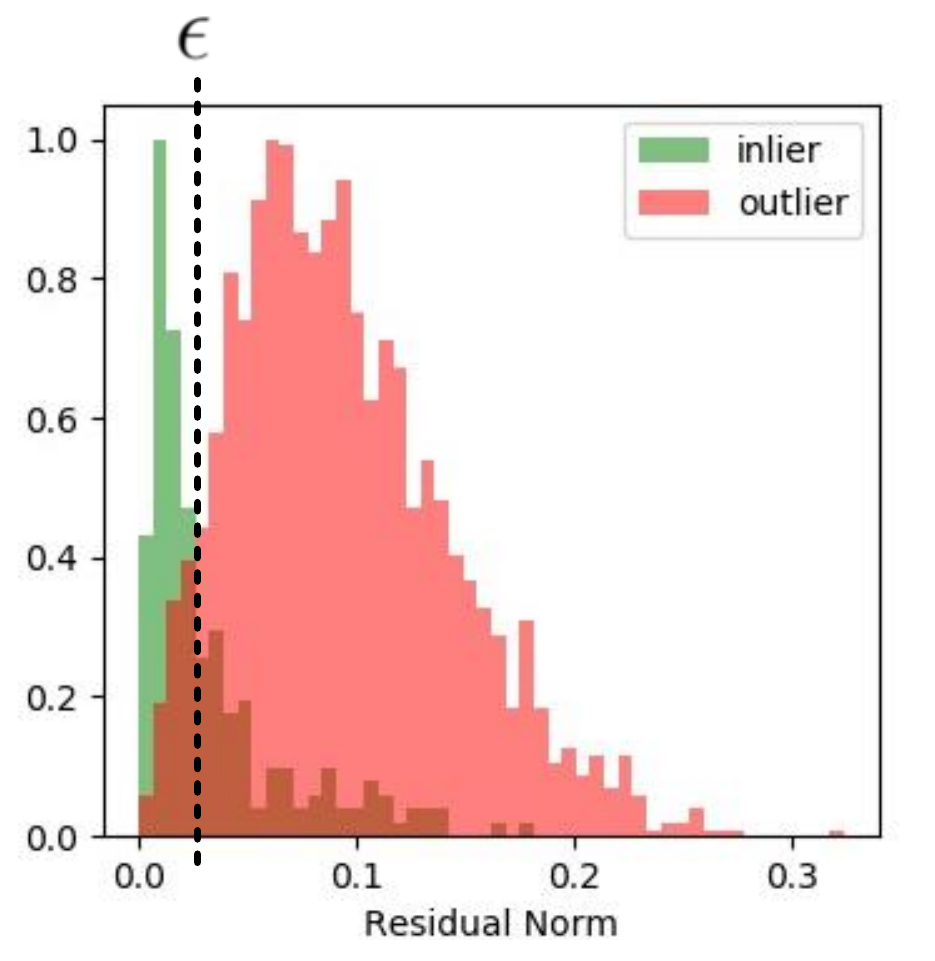} \\
    (a) Input correspondences & (b) Output of LMF & (c) Histogram of residuals
    \end{tabular}
    \caption{(a) 2000 putative correspondences. Green correspondences are true while red ones are false.  (b) Output correspondences of LMF. (c) Distribution of motion residual norms of inliers ({\color{green} green}) and outliers ({\color{red} red}). Histograms are normalized with 1 as the max values. }
    \label{fig:gcf}
\end{figure*}

Denote $\bm{R}(\eta) \equiv \bm{U} diag([1/(1+\eta\lambda_i)]) \bm{U}^\intercal$. Then, the residuals between the smoothed motions and the input motions are computed by $\bm{R}(\eta)\bm{m}-\bm{m}$. Since only true correspondences can be well fitted by the smooth motions $\bm{R}(\eta)\bm{m}$ while false correspondences cannot, the residuals of true correspondences will be significantly smaller than those of false correspondences. Hence, the true correspondences can be distinguished from the false ones by thresholding on the L2-norms of the residual motions. The whole process, called {\em Laplacian Motion Fitting} (LMF), is summarized in Algorithm~\ref{alg:gcf}, and an example is shown in Fig.~\ref{fig:gcf}.

\begin{algorithm}
    \SetAlgoLined
    \KwData{Input correspondences $\{\bm{c}_i=(x_i,y_i,u_i,v_i)\}$, smooth strength $\eta$ and inlier threshold $\epsilon$}
    \KwResult{Probability of being inliers $\{p_i\}$ }
    Compute Laplacian matrix $\bm{L}$ on $\{\bm{c}_i\}$\;
    Eigen decomposition of $\bm{L}=\bm{U} \bm{\Lambda} \bm{U}^\intercal$\;
    Compute motions $\bm{m}=[\bm{m}_i]=[(u_i-x_i,v_i-y_i)]$\;
    Compute smoothed motions $\bm{s}=[\bm{s}_i]=\bm{R}(\eta)\bm{m}$\;
    Find inliers $\{p_i=1\ if\ \|\bm{s}_i-\bm{m}_i\|_2\le \epsilon\ else\ 0\}$\;
    \caption{Laplacian Motion Fitting}
    \label{alg:gcf}
\end{algorithm}

\textbf{Computational complexity}. For a graph Laplacian matrix, its eigenvalues $\lambda_i$ are non-negative.  As $\lambda_i$ increases, the $1/(1+\eta \lambda_i)$ becomes sufficiently small and so negligible. In light of this, we can just use the $k_e$ smallest eigenvalues and their associated eigenvectors for the computation of $\bm{R}(\eta)$, which means changing $\bm{U}\in\mathbb{R}^{N\times N}$ to $\mathbb{R}^{N\times k_e}$.
This simplification essentially lowers the computational complexity of $\bm{R}(\eta)$ from $O(n^2)$ to $O(n)$.

\textbf{Graph construction.} 
In our implementation, the graph is constructed by connecting every correspondence with its $k$-nearest neighbors in the coordinate space $\bm{c}_i\in \mathbb{R}^4$, which is the ``bilateral space" proposed in CODE~\cite{lin2017code}. This bilateral space allows finding a piece-wise smooth motion field rather than a global smooth motion field. Meanwhile, the graph construction can be quite flexible. We can also adopt the affine compatibility proposed in \cite{zhao2019nm} for graph construction to utilize detected affine transformations. 

\subsection{Coherence Residual Layer}
\label{sec:crl}
So far, we have presented the LMF algorithm on motions for correspondences pruning. However, due to the sparsity and uneven distributions of putative correspondences, expert knowledge is needed for carefully tuning the threshold $\epsilon$ and the smoothness strength $\eta$ in order to achieve a better performance. To avoid this, we incorporate the LMF within a learning framework so that we can utilize the powerful representational ability of neural networks to automatically learn complex motion patterns from training data.

Instead of directly thresholding on residual motions, we apply the LMF on the features $\bm{f}_l\in \mathbb{R}^{N\times d}$ of correspondences, which are extracted by a neural network, to find a set of smoothed features $\bm{f}'_l \in \mathbb{R}^{N\times d}$. Then, discriminative features can be extracted from the residuals $\bm{f}_l-\bm{f}'_l$ for the classification of correspondences.
We consider the problem of estimating smooth features by,
\begin{equation}
    \mathop{minimize}\limits_{\bm{f}'_l}^{}\  \Tr((\bm{f}'_l-\bm{f}_l)^\intercal(\bm{f}'_l-\bm{f}_l))+\eta \Tr(\bm{f}_l^{\prime\intercal}\bm{L}\bm{f}'_l),
    \label{eq:obj_feats}
\end{equation}
which is similar to Problem (\ref{eq:obj_vec}) by replacing the input motions $\bm{m}$ with the features $\bm{f}_l$. 
However, the features $\bm{f}_l$ are extracted by the neural network via multiple layers of abstractions and thus Problem~(\ref{eq:obj_feats}) is much more generalized than fitting a single smooth motion field in Problem~(\ref{eq:obj_vec}). 
These features can simply be motions if necessary or any other more complex coherent attributes like local affine transformations~\cite{lin2014bilateral,lin2011smoothly}, which are implicitly learned by the neural network during training.

The solution to Problem~(\ref{eq:obj_feats}) is also given by $\bm{f}'_l=\bm{R}(\eta)\bm{f}_l$ according to Proposition~\ref{prop:solution}. Based on this, we formulate a new layer called {\em Coherence Residual Layer} (CR-Layer) by,
\begin{equation}
    \bm{f}_{l+1}={\rm ContextNorm}(\bm{f}_{l}-\bm{R}(\eta) \bm{f}_{l}),
\end{equation}
where the $\bm{f}_{l+1}$ are output features and ContextNorm~\cite{moo2018learning} contains a Multi-Layer Percetron (MLP) and an instance normalization layer for feature extraction. The forward pass of CR-Layer implicitly solves Problem~(\ref{eq:obj_feats}) by $\bm{R}(\eta) \bm{f}_{l}$ and extract features from the residuals $\bm{f}_{l}-\bm{R}(\eta) \bm{f}_{l}$. 
Since the whole process only involves matrix multiplication, CR-Layer is differentiable and therefore can be incorporated in a network. Meanwhile, We also learn the smoothness strength $\eta$ from data by making it a trainable parameter.

\subsection{Local Coherence Layer}
\label{sec:lgc}
\begin{figure}
    \centering
    \includegraphics[width=0.95\linewidth]{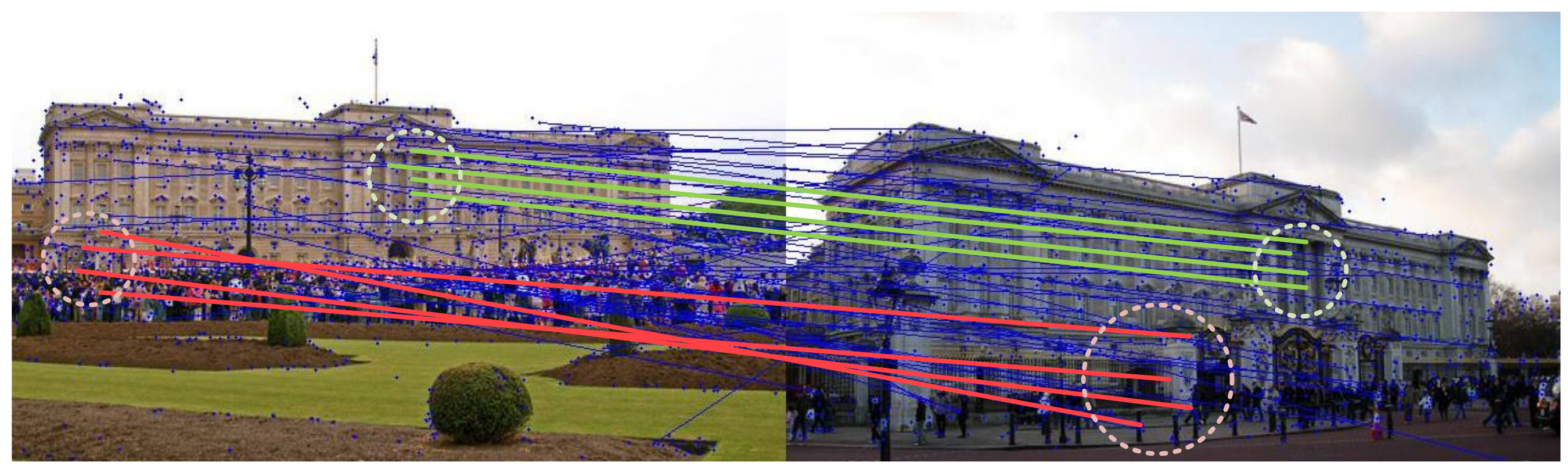}
    \caption{True correspondences ({\color{green}green}) have motion-consistent supporting neighbors while neighbors of false correspondences ({\color{red}red}) scatter randomly.}
    \label{fig:lgc}
\end{figure}

Besides fitting a global smooth function on correspondences, another important observation of motion coherence is that true correspondences tend to have motion-consistent supporting correspondences in their neighborhoods, while false correspondences do not have such supporting correspondences. A typical example is shown in Fig.~\ref{fig:lgc}.
Based on this observation, we introduce a new layer called {\em Local Coherence Layer} (LC-Layer).

Specifically, given the feature $\bm{f}_{l,i}$ of the $i$-th correspondence, we first compute the feature differences between the correspondence and its neighbors $(i,j)\in E$ by $\bm{f}_{l,i}-\bm{f}_{l,j}$. The feature differences measure the local consistency in the neighborhood of the $i$-th correspondence. Then, the LC-Layer is defined by,
\begin{equation}
  \bm{f}_{l+1,i}=\mathop{{\rm MaxPool}}\limits_{(i,j)\in E} \left(\{ {\rm MLP}(\bm{f}_{l,i}-\bm{f}_{l,j})\} \right),
  \label{eq:lgc}
\end{equation}
where $\rm MLP$ is a multi-layer perceptron, $\rm MaxPool$ is a max-pooling operator which pools on all neighboring features to get a single feature vector, and $\bm{f}_{l+1,i}$ is the output feature of this correspondence which contains information about the local coherence in its neighborhood.

\subsection{Architecture}
\label{sec:arc}
\begin{figure*}
    \centering
    \includegraphics[width=0.95\textwidth]{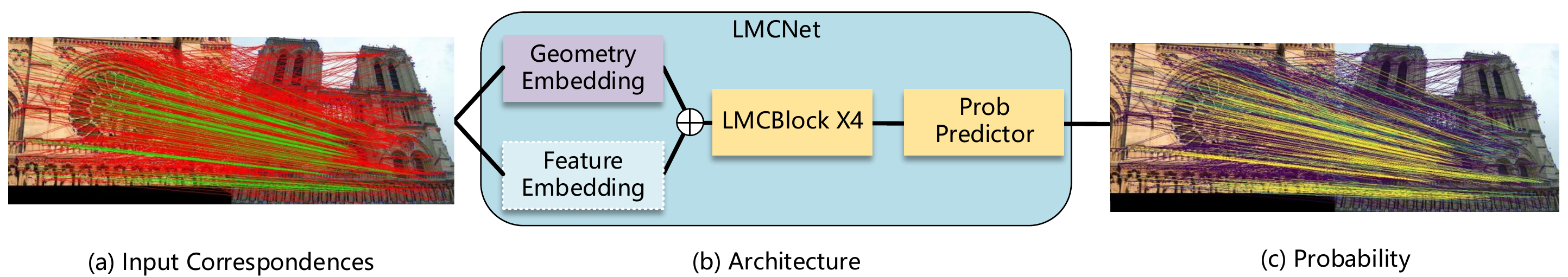} 
    \caption{(a) Input correspondences. Red represents false correspondences while green represents true ones. (b) Architecture of LMCNet. The feature embedding is optional. (c) Output probability of being inliers. Brighter color means higher probability. }
    \label{fig:arch}
\end{figure*}
\textbf{Overview}. The architecture of our network is illustrated in Fig.~\ref{fig:arch}. Given the input correspondences, the Geometry Embedding is a ContextNorm layer which processes the input coordinates $\bm{c}\in \mathbb{R}^{N\times 4}$ to produce $d$-dimensional features $\bm{f}_1\in \mathbb{R}^{N\times d}$. If there are optional features associated with correspondences, we process these features with another ContextNorm layer (Feature Embedding), and we add it back to the output of Geometry Embedding.  Then, the features are processed by 4 blocks called LMCBlocks, which are the main feature extraction module of LMCNet. Finally, the output features of LMCBlocks $\bm{f}_{out}\in \mathbb{R}^{N\times d}$ are processed by a probability predictor, which simply consists of a fully connected layer and a sigmoid function, to produce the inlier probability $\bm{p}=[p_i]\in \mathbb{R}^N$.

\begin{figure}
    \centering
    \includegraphics[width=0.45\textwidth]{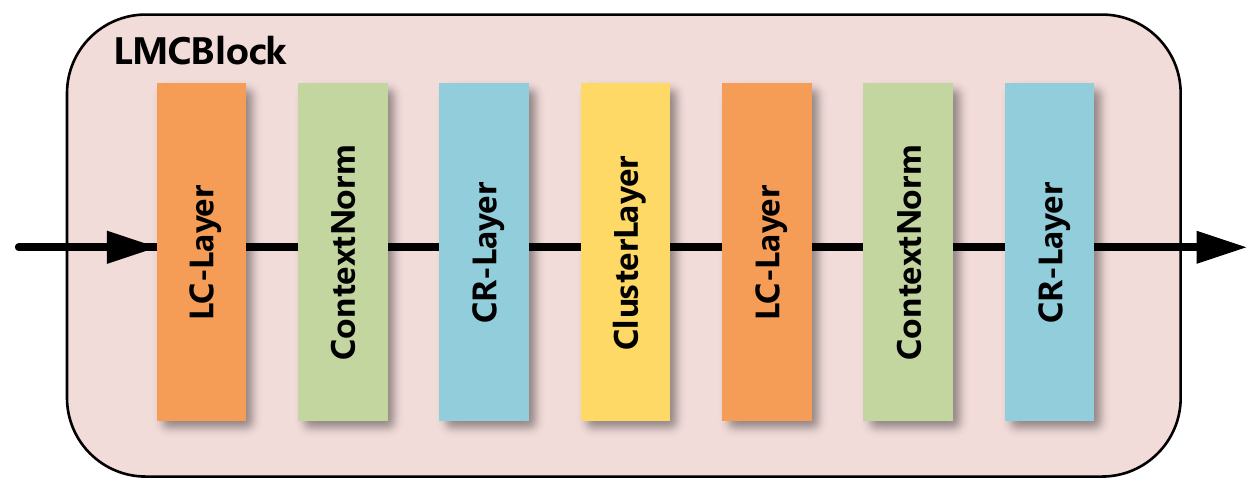} 
    \caption{A LMCBlock consists of 7 layers.}
    \label{fig:block}
\end{figure}
\textbf{LMCBlock}. The structure of a LMCBlock is illustrated in Fig.~\ref{fig:block}. The LC-Layer is placed on the top to extract some useful information from the neighborhoods of correspondences and the CR-Layer is placed after several other layers so that its inputs are at a higher level of abstraction thus more flexible. Except for the proposed LC-Layers and CR-Layers, a LMCBlock also includes two ContextNorm layers and a clustering layer, which are used for extracting other information such as the underlying epipolar geometry. The clustering layer used here is proposed in~\cite{zhang2019learning}, which is implemented by differentiable pooling to $k_c$ clusters, order-aware filtering among clusters and unpooling to original correspondences. All layers in the LMCBlock take $d$-dimensional features as inputs and also output $d$-dimensional features. Thus, a skip connection is applied to add the inputs to the output features on all layers. We also use a bottleneck structure in LC-Layers, which encodes the input features into a lower dimension $d_l$, then extracts the local consistency features on the low-dimensional features, and finally lift the dimension back to $d$.

\subsection{Implementation Details}
\label{sec:impl}

In our implementation, the coordinates of correspondences are normalized to the range $[-1,1]$ on dynamic scene datasets or by camera intrinsic matrices in the relative pose estimation. For the construction of the adjacency matrix, we use $k=8$ neighbors and $\sigma=0.1$. We use the normalized graph Laplacian matrix $\hat{\bm{L}}=\bm{D}^{-1/2}\bm{L}\bm{D}^{-1/2}$ to compute $\bm{R}(\eta)$ and only the smallest $k_e=32$ eigenvalues and the associated eigenvectors are selected. In all the CR-Layers, $\eta$ is initialized to 10.0. Both the feature dimension $d$ and the cluster number $n_c$ in LMCBlock are $128$, and the bottleneck feature dimension $d_l$ used in LC-Layers is 8. A correspondence is determined as a inlier if its predicted inlier probability is larger than 0.95. More details about the architecture and the training process can be found in the supplementary material. When implementing the LMF algorithm, we use $k_e=128$ eigenvalues, the smoothness strength $\eta=10.0$ and the threshold $\epsilon=0.025$.

\textbf{Loss}. For image pairs of dynamic scenes, we use the binary cross entropy loss $\ell_{cls}$ for training. For image pairs in relative pose estimation, we use an additional geometric loss $\ell_{geom}$~\cite{zhang2019learning,ranftl2018deep,hartley2003multiple}, in which we estimate the essential matrix by the weighted 8 points algorithm~\cite{hartley2003multiple} and compute distances from ground-truth correspondences to estimated epipolar lines as loss.

\section{Experiments}

\subsection{Evaluation protocols}
To demonstrate the effectiveness of our methods, we evaluate three models, which are the LMF algorithm, the LMCNet with only coordinates as input and the LMCNet with both coordinates and local descriptors as inputs (LMCNet-F). We report performances on relative pose estimation and correspondence pruning of dynamic scenes.

\textbf{Relative pose datasets}. The outdoor YFCC100M~\cite{thomee2016yfcc100m} dataset and the indoor SUN3D~\cite{xiao2013sun3d} dataset are used for relative pose estimation. We use the same train-test split as~\cite{zhang2019learning}. Input putative correspondences are generated from nearest neighborhood matching of 2000 SIFT~\cite{Lowe2004Distinctive} descriptors on every image. 
We regard correspondences with small distances ($\le 0.01$ in normalized image coordinate) to their ground-truth epipolar lines as true correspondences. 
For a image pair, an essential matrix is estimated by RANSAC on the predicted true correspondences and is then decomposed to a rotation and a translation.

\textbf{Dynamic scene datasets}.
The dynamic scene dataset contains images with dynamic objects. We use the DETRAC dataset~\cite{detrac} for evaluation. The DETRAC dataset contains images in traffic monitoring, so the background is fixed and the main dynamic objects are cars. We extract 2048 SuperPoint~\cite{detone2018superpoint} features on every image and extract putative correspondences by nearest neighborhood matching. 
Since there are only annotated bounding boxes of different car instances in the dataset, we regard a correspondence as a true correspondence if it connects two bounding boxes of the same instance while a correspondence is false if it connects different instances or it connects an instance with background, as shown in Fig.~\ref{fig:detrac}. 
We use the provided train-test split with 60 sequences (sample 30k image pairs) for training and 40 sequences (sample 4k image pairs) for testing.
We also include some qualitative results on the DAVIS~\cite{davis2017} dataset in the supplementary material.

\textbf{Metrics}. On the task of relative pose estimation, we compute the Area Under the Curve (AUC) of the pose accuracy curve at thresholds $5^{\circ}$, $10^{\circ}$ and $20^{\circ}$, the same as used in~\cite{sarlin20superglue}.
On the task of correspondence pruning on dynamic image pairs, we report the precision, recall and F1 scores.

\subsection{Results on relative pose estimation}
\newcommand\ph{0.085}
\newcommand\pose{pose_results_low}
\begin{figure*}
\centering
\setlength\tabcolsep{1pt}
\begin{tabular}{m{1.5cm} m{4.5cm} m{3.2cm} m{3.1cm} m{3.5cm}}
Inputs &
\includegraphics[height=\ph\textwidth]{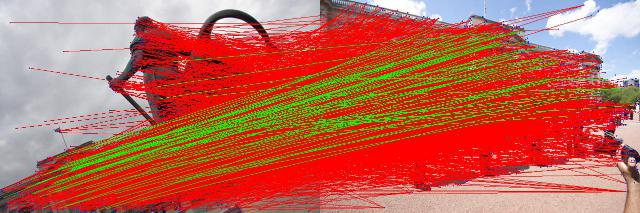} &
\includegraphics[height=\ph\textwidth]{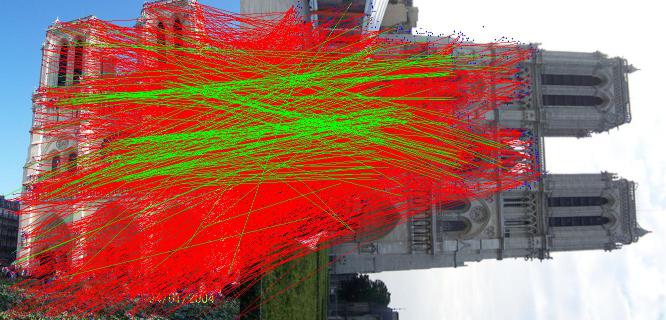} &
\includegraphics[height=\ph\textwidth]{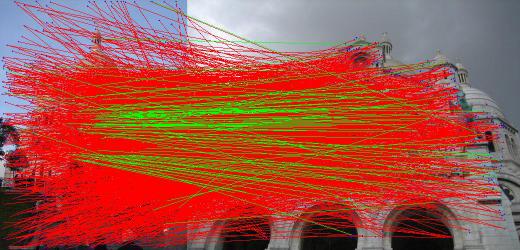} &
\includegraphics[height=\ph\textwidth]{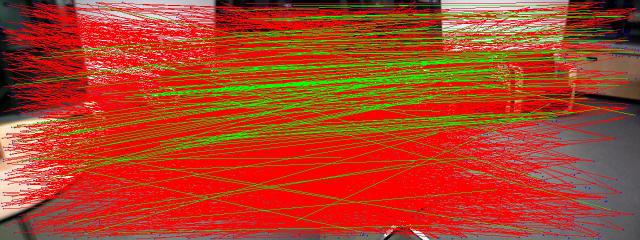} \\
PointCN &
\includegraphics[height=\ph\textwidth]{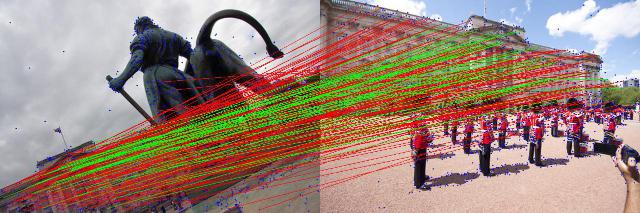} &
\includegraphics[height=\ph\textwidth]{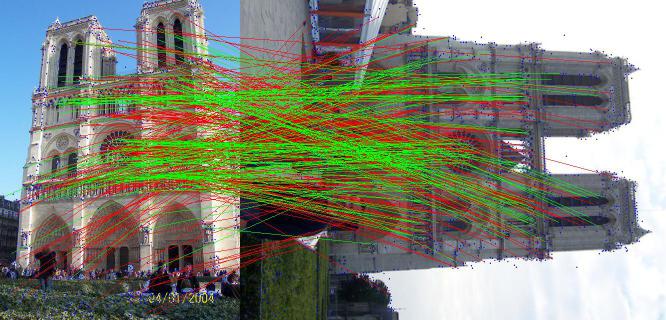} &
\includegraphics[height=\ph\textwidth]{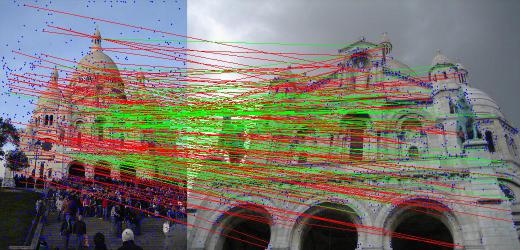} &
\includegraphics[height=\ph\textwidth]{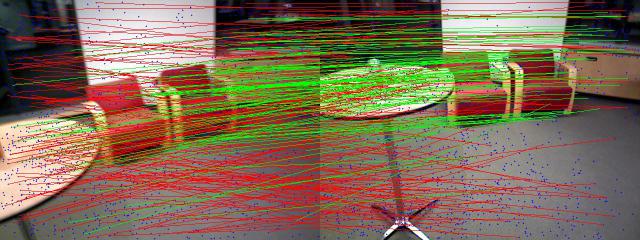} \\
OANet &
\includegraphics[height=\ph\textwidth]{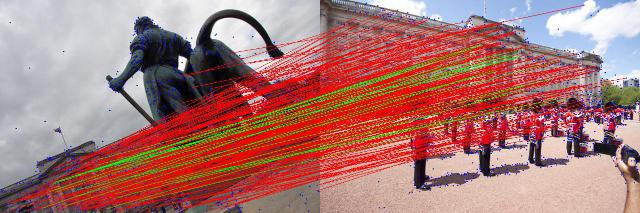} &
\includegraphics[height=\ph\textwidth]{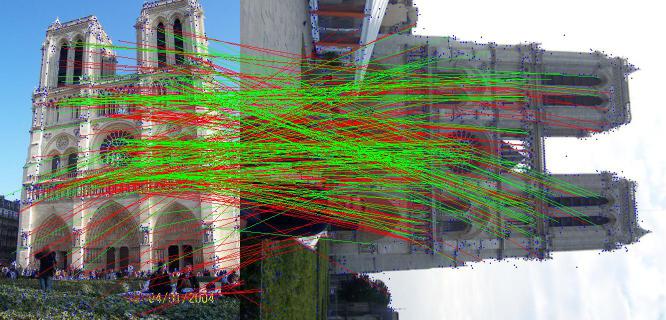} &
\includegraphics[height=\ph\textwidth]{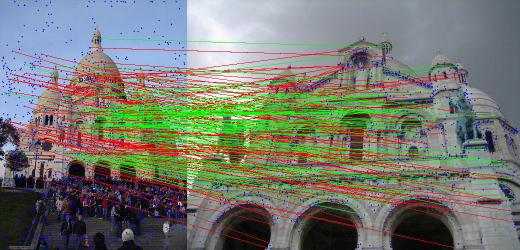} &
\includegraphics[height=\ph\textwidth]{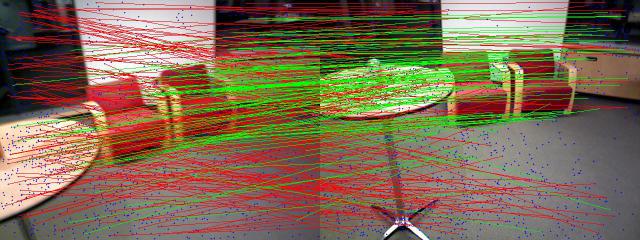} \\
LMCNet &
\includegraphics[height=\ph\textwidth]{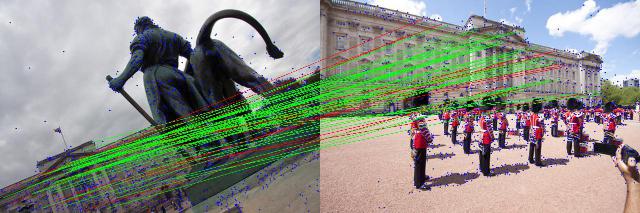} &
\includegraphics[height=\ph\textwidth]{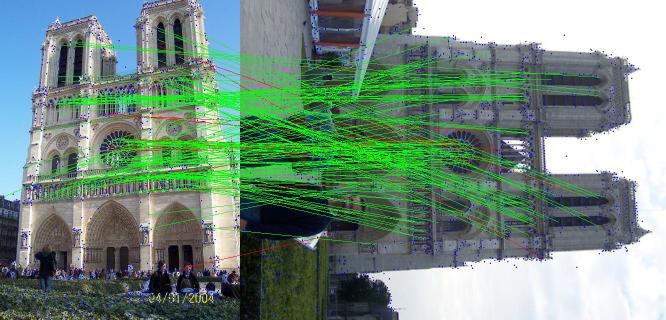} &
\includegraphics[height=\ph\textwidth]{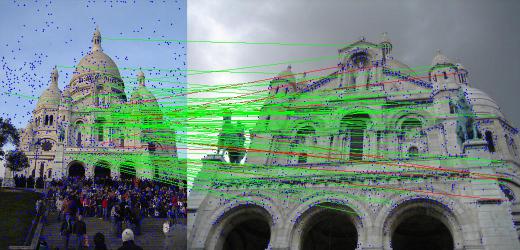} &
\includegraphics[height=\ph\textwidth]{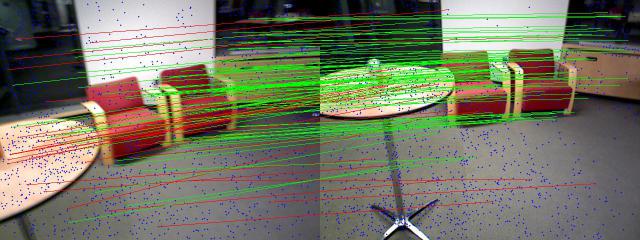}
\end{tabular}
\caption{Input correspondences (Row 1), output of  PointCN~\cite{moo2018learning} (Row 2), OANet~\cite{zhang2019learning} (Row 3) and LMCNet (Row 4). Green correspondences are correct (small distances to true epipolar lines) while red correspondences are incorrect.}
\label{fig:pose}
\end{figure*}
\newcommand{\cmark}{\ding{51}}%
\newcommand{\xmark}{\ding{55}}%
\begin{table}
  \centering
    \setlength{\tabcolsep}{1mm}{
    \begin{tabular}{lccccc}
    \toprule
    Method & Desc & AUC@$5^{\circ}$ & AUC$10^{\circ}$ & 
    AUC$20^\circ$ \\
    \hline
    Ratio test & \cmark & 24.09 & 40.71 & 58.14 \\
    LPM~\cite{ma2019locality} & \xmark & 10.48 & 18.91 & 29.26 \\
    GMS~\cite{bian2017gms} & \xmark & 19.05 & 32.35 & 46.79 \\
    CODE~\cite{lin2017code} & \xmark & 16.99 & 30.23 & 43.85 \\
    LMF & \xmark & 16.91 & 29.49 & 43.44 \\
    \hline
    PointCN~\cite{moo2018learning} & \xmark & 27.39 & 44.61 & 61.22 \\
    AttenCN~\cite{sun2020acne} & \xmark & 29.08 & 48.13 & 65.49 \\
    OANet~\cite{zhang2019learning} & \xmark & 29.12 & 48.28 & 65.37 \\
    SuperGlue~\cite{sarlin20superglue} & \cmark & 30.49 & 51.29 & 69.72 \\
    LMCNet & \xmark & 34.62 & 53.86 & 70.53 \\
    LMCNet-F & \cmark & \textbf{35.9}1 & \textbf{55.68} & \textbf{72.35} \\
    \bottomrule
    \end{tabular}%
    }
  \vspace{5pt}
  \caption{Pose AUCs of LMCNet and other baseline methods on the YFCC100M dataset. The column ``Desc" means the model uses descriptors as inputs or not. If not, the model takes only the coordinates of correspondences as inputs.}
  \label{tab:outdoor}%
\end{table}%
\begin{table}
  \centering
    \setlength{\tabcolsep}{1mm}{
    \begin{tabular}{lccccc}
    \toprule
    Method & AUC@$5^{\circ}$ & AUC$10^{\circ}$ & 
    AUC$20^\circ$ \\
    \hline
    Ratio test & 4.51 & 11.62 & 23.02 \\
    LPM~\cite{ma2019locality} & 2.81 & 7.40 & 15.36 \\
    GMS~\cite{bian2017gms} & 4.36 & 11.08 & 21.68 \\
    CODE~\cite{lin2017code} & 3.52 & 8.91 & 18.32 \\
    LMF & 3.34 & 8.85 & 18.04 \\
    \hline
    PointCN~\cite{moo2018learning} & 5.64 & 14.88 & 29.32 \\
    AttenCN~\cite{sun2020acne} & 5.97 & 15.69 & 30.98 \\
    OANet~\cite{zhang2019learning} & 5.94 & 15.79 & 31.03 \\
    LMCNet & 6.77 & 17.14 & 32.55 \\
    LMCNet-F & \textbf{8.86} & \textbf{19.64} & \textbf{34.96} \\
    \bottomrule
    \end{tabular}%
    }
  \vspace{5pt}
  \caption{Pose AUCs of LMCNet and other baseline methods on the indoor SUN3D dataset.}
  \label{tab:indoor}%
\end{table}%
\textbf{Baselines}. We consider both the traditional handcrafted pruners, including LPM~\cite{ma2019locality}, GMS~\cite{bian2017gms} and CODE~\cite{lin2017code}, and learning-based pruners, including PointCN~\cite{moo2018learning}, AttenCN~\cite{sun2020acne} and OANet~\cite{zhang2019learning} as baseline methods. We also include the results of the learning-based matcher SuperGlue~\cite{sarlin20superglue} on the YFCC100M dataset. For implementation, we directly use the released code of GMS~\cite{bian2017gms} and LPM~\cite{ma2019locality}, and we re-implement CODE~\cite{lin2017code} which takes only coordinates of correspondences as inputs. For learning-based methods, we directly use the released code and these methods are trained on the same training set with the same learning rates, except for SuperGlue~\cite{sarlin20superglue}, of which we directly use the results reported in their supplementary material~\cite{sarlin20superglue}.

\textbf{Results}. The quantitative results are reported in Table~\ref{tab:outdoor} and Table~\ref{tab:indoor}. Some qualitative results are provided in Fig.~\ref{fig:pose}. The results show that LMCNet outperforms all the baseline methods on all pose metrics. Further adding descriptors as inputs (LMCNet-F) leads to about 1.5\%-2\% improvements on both datasets. 
Among the handcrafting methods, the results of LMF are similar to the global method CODE~\cite{lin2017code} but are inferior to the local method GMS~\cite{bian2017gms}. This is due to the unevenly-distributed correspondences which make it hard for global methods to determine a uniform global threshold for pruning. However, learning the motion coherence via CR-Layers enables the network to handle complex motion patterns and thus achieves a better performance.

\subsection{Results on dynamic scenes}
\begin{table}
  \centering
    \setlength{\tabcolsep}{1mm}{
    \begin{tabular}{lcccc}
    \toprule
    Method &  Precision & Recall & F1-Score\\
    \hline
    LPM~\cite{ma2019locality} & 85.12 & 52.27 & 63.47 \\
    GMS~\cite{bian2017gms}  & 84.89 & 76.92 & 80.01 \\
    CODE~\cite{lin2017code} & 83.63 & 77.86 & 79.95 \\
    LMF & 82.84  & 78.66 & 79.35 \\
    \hline
    PointCN~\cite{moo2018learning} & 84.27 & 89.34 & 86.26 \\
    AttenCN~\cite{sun2020acne} & 85.34 & 88.53 & 86.89 \\
    OANet~\cite{zhang2019learning} & 82.70 & 87.66 & 84.53 \\
    LMCNet & \textbf{87.23} & \textbf{91.16} & \textbf{88.79} \\
    \bottomrule
    \end{tabular}%
    }
  \vspace{5pt}
  \caption{Precision, recall and F1 score of LMCNet and other baseline models on the DETRAC dataset.}
  \label{tab:detrac}%
  
\end{table}%
\newcommand\phnew{0.08}
\newcommand\detrac{detrac_low}
\begin{figure*}
\centering
\setlength\tabcolsep{1pt}
\begin{tabular}{m{1.5cm} m{5cm} m{5cm} m{5cm}}
Inputs &
\includegraphics[height=\phnew\textwidth]{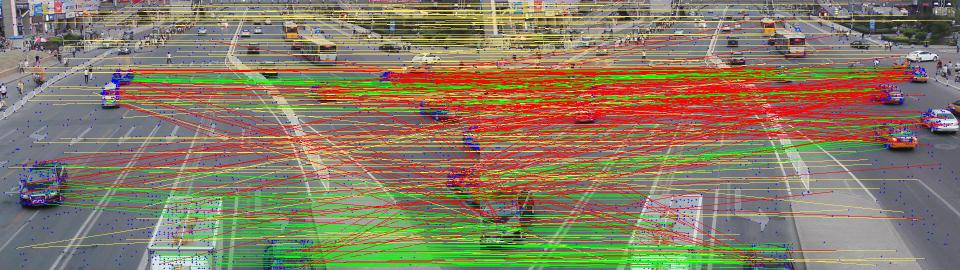} &
\includegraphics[height=\phnew\textwidth]{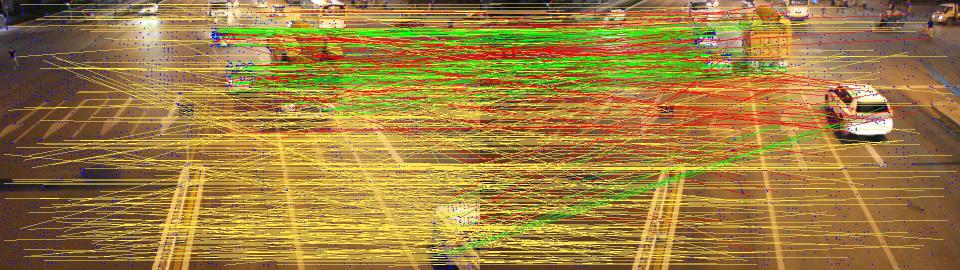} &
\includegraphics[height=\phnew\textwidth]{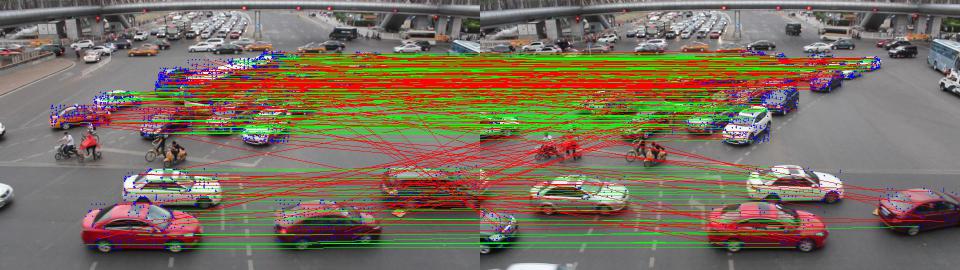} \\
PointCN &
\includegraphics[height=\phnew\textwidth]{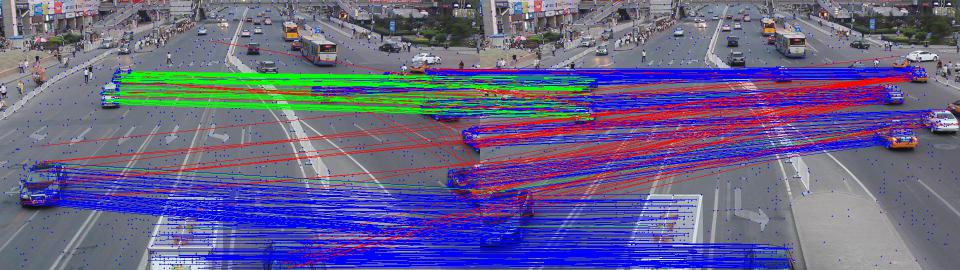} &
\includegraphics[height=\phnew\textwidth]{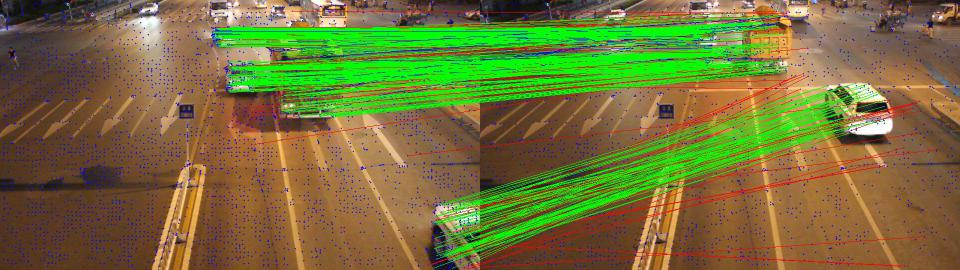} &
\includegraphics[height=\phnew\textwidth]{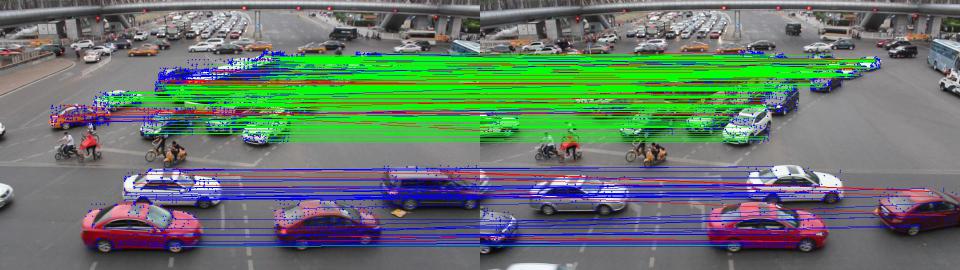} \\
OANet &
\includegraphics[height=\phnew\textwidth]{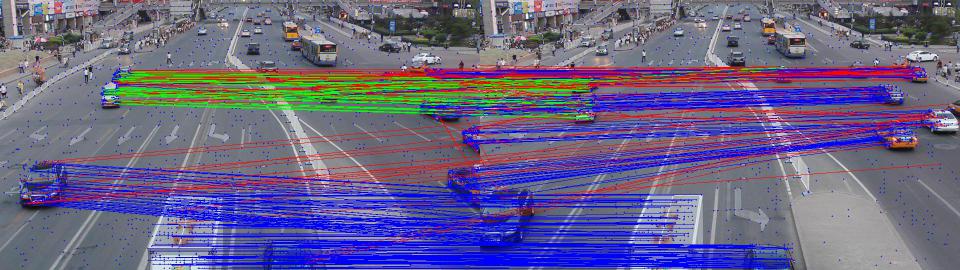} &
\includegraphics[height=\phnew\textwidth]{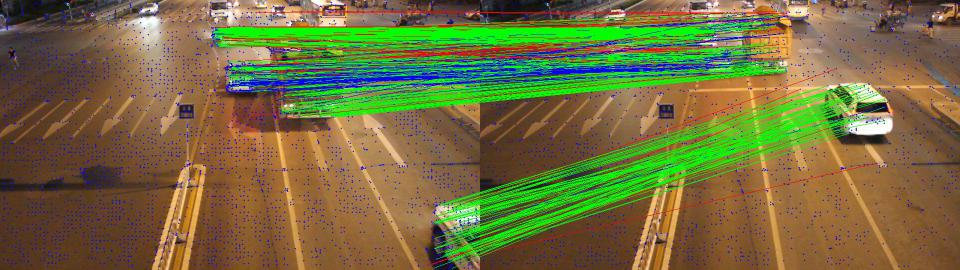} &
\includegraphics[height=\phnew\textwidth]{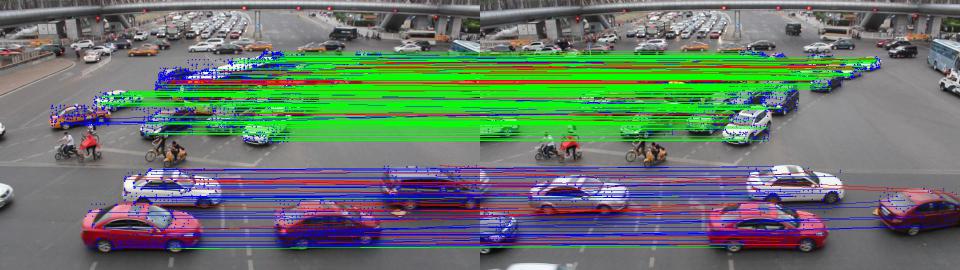} \\
LMCNet &
\includegraphics[height=\phnew\textwidth]{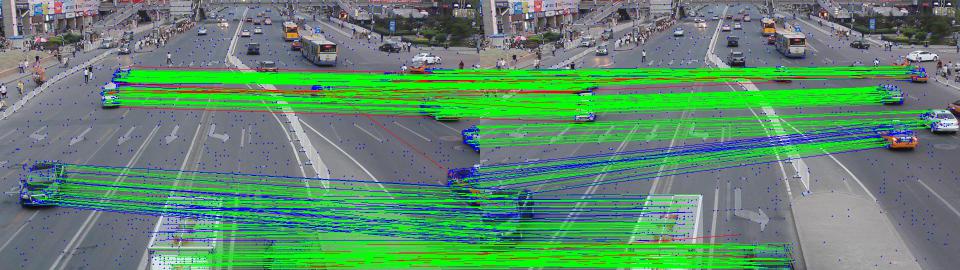} &
\includegraphics[height=\phnew\textwidth]{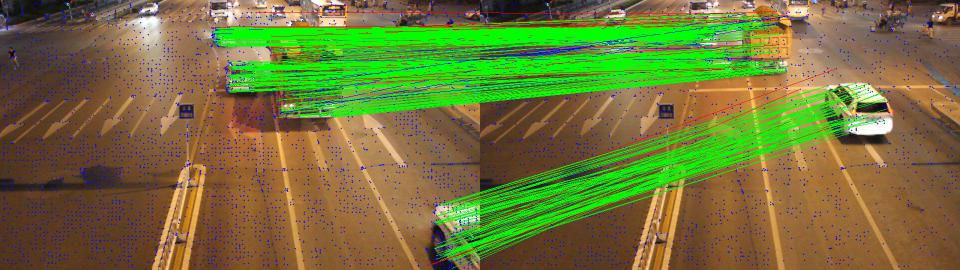} &
\includegraphics[height=\phnew\textwidth]{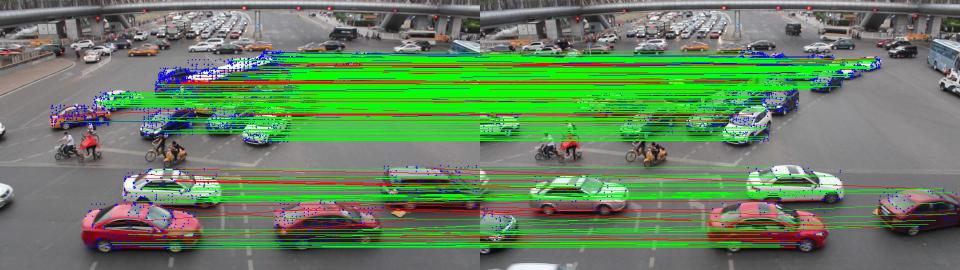} \\
\end{tabular}
\caption{Input correspondences (Row 1), output correspondences of PointCN~\cite{moo2018learning} (Row 2), OANet~\cite{zhang2019learning} (Row 3) and LMCNet (Row 4) on the DETRAC dataset. Yellow means that correspondences are in the background, Green means correct, Red means false positive and blue color means false negative. 
Note the inputs include both foreground and background correspondences. For clear visualization, we randomly draw 512 correspondences on all image pairs and neglect the background correspondences in Row 2, 3 and 4.}
\label{fig:detrac}
\end{figure*}
On the DETRAC dataset, we compare LMCNet with the same baseline methods as used for relative pose estimation. The quantitative results in Table~\ref{tab:detrac} show that LMCNet outperforms all the baseline methods in all the metrics. From qualitative results in Fig.~\ref{fig:detrac}, we can see that LMCNet is able to robustly find motion-coherent correspondences while baseline methods may include some false correspondences or neglect some sparse true correspondences.

\subsection{Analysis}
\begin{table}
  \centering
    \setlength{\tabcolsep}{1mm}{
    \begin{tabular}{ccccccc}
    \toprule
    Base- & CR-   & LC-   & $k_e$ & AUC@ & AUC@ & AUC@ \\
    line  & Layer & Layer &       & $5^\circ$ & $10^\circ$ & $20^\circ$ \\
    \hline
    \cmark &        &        & 32 & 29.38 & 48.00 & 64.86 \\
    \cmark & \cmark &        & 32 & 32.72 & 51.66 & 69.49 \\
    \cmark & \cmark & \cmark & 32 & 34.62 & 53.86 & 70.53 \\
    \hline
    \cmark & \cmark & \cmark & 16 & 32.58 & 51.72 & 69.51 \\
    \cmark & \cmark & \cmark & 32 & 34.62 & 53.86 & 70.53 \\
    \cmark & \cmark & \cmark & 64 & 34.71 & 53.79 & 70.63 \\
    \bottomrule
    \end{tabular}%
    }
  \vspace{5pt}
  \caption{Ablation studies of LMCNet on the YFCC100M dataset. The baseline model has the same structure as LMCNet but replaces all CR-Layers and LC-Layers with ContextNorm layers.}
  \label{tab:ablation}%
\end{table}%
\textbf{Ablation studies}.
To demonstrate the effectiveness of LC-Layers and CR-Layers, we perform ablation studies on the YFCC100M dataset and report the results in Table~\ref{tab:ablation}. The baseline model replaces all CR-Layers and LC-Layers with ContextNorm layers with almost equivalent computational complexity and parameter numbers. The results show that adding CR-Layers brings about 2-5\% improvements to all pose AUCs, and further adding LC-Layers results in 1-2\% improvements.

\textbf{Different numbers of eigenvectors in CR-Layers}.
As mentioned in Sec.~\ref{sec:gcf}, we can use $k_e$ smallest eigenvalues and their associated eigenvectors in the computation of CR-Layers. To show how the number of eigenvectors $k_e$ affects performances, we train LMCNet with different $k_e$ and report the results in Table~\ref{tab:ablation}. The results show that using only 16 eigenvectors decreases the performances remarkably compared to using 32 eigenvectors. However, increasing the number from 32 to 64 does not bring about significant performance improvement. In light of this, we use 32 eigenvectors in LMCNet for computational efficiency.

\begin{figure}
    \centering
    \setlength\tabcolsep{1pt}
    \begin{tabular}{cc}
    \includegraphics[width=0.24\textwidth]{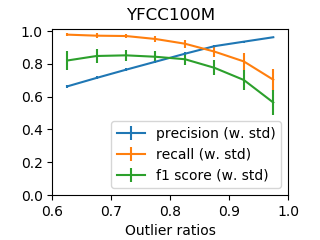} &
    \includegraphics[width=0.24\textwidth]{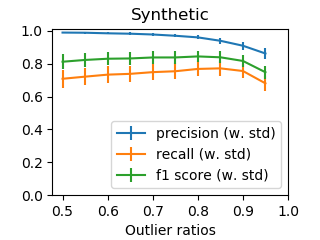} \\
    (a)  & (b)
    \end{tabular}
    \caption{Precision, recall and F1 scores of LMCNet on the YFCC100M dataset (a) and the synthetic dataset (b). 
    }
    \label{fig:outlier}
\end{figure}
\textbf{Robustness to outliers}.
To show the robustness of LMCNet to outlier ratios, we report performances of LMCNet under different outlier ratios. In Fig.~\ref{fig:outlier} (a), we compute the precision, recall and F1 scores in different outlier ratio ranges on the YFCC100M dataset. 
Surprisingly, as the outlier ratio increases, the recall and the F1 scores decrease but the precision increases. Because in this experiment, the total number of correspondences is fixed so that increasing outlier ratio means fewer inliers. If there are fewer inliers, LMCNet will also predict fewer but more confident inliers accordingly. We further conduct a synthetic experiment by manually adding random outliers to 100 fixed image pairs in the YFCC100M, as shown in Fig.~\ref{fig:outlier} (b). In this experiment, the performance of LMCNet drops slightly, which also demonstrates its robustness to outliers.

\textbf{Compatibility with learning-based descriptors and matchers}.
In Table~\ref{tab:advance}, we report performances with or without LMCNet on the SUN3D dataset using SuperPoint~\cite{detone2018superpoint} as the local descriptor and SuperGlue~\cite{sarlin20superglue} as the matcher. In this experiment, SuperGlue and SuperPoint only uses 1024 keypoints but still achieve a better performance than using 2000 SIFT features. Meanwhile, applying LMCNet as a pruner can further improve the accuracy of estimated poses in all cases. Qualitative results and details about this experiment can be found in the supplementary material. 

\begin{table}
  \centering
    \setlength{\tabcolsep}{1mm}{
    \begin{tabular}{cccccc}
    \toprule
    \multirow{2}{*}{Desc} & \multirow{2}{*}{Matcher} & \multirow{2}{*}{Pruner} & AUC@ & AUC@ & AUC@ \\
     & & & 5$^\circ$ & 10$^\circ$ & 20$^\circ$ \\
    \hline
    SP   & NN &    /   & 4.66 & 12.49 & 25.39 \\
    SP   & NN & LMCNet & 6.76 & 17.25 & 32.90 \\
    \hline
    SP   & SG &    /   & 7.09 & 17.82 & 33.26 \\
    SP   & SG & LMCNet & 8.13 & 20.36 & 37.55 \\
    \bottomrule
    \end{tabular}%
    }
  \vspace{5pt}
  \caption{Pose AUCs of LMCNet on the indoor SUN3D dataset with the SuperPoint (SP)~\cite{detone2018superpoint} descriptor and the SuperGlue (SG)~\cite{sarlin20superglue} matcher. NN means the nearest neighbor matcher.}
  \label{tab:advance}%
\end{table}%

\section{Conclusion}
In this paper, we designed a novel architecture LMCNet to learn the motion coherence property for correspondence pruning. We proposed a novel formulation LMF of the motion coherence by fitting a smooth function via decomposition of graph Laplacian, which enables us to design a differentiable CR-Layer to capture the global motion coherence in a neural network. Furthermore, we also designed a LC-Layer to extract local coherence information from neighborhoods of correspondences. Integrating these two coherence layers, the proposed LMCNet achieves superior performances on relative pose estimation and correspondence pruning of dynamic scenes.

{\small
\bibliographystyle{ieee_fullname}
\bibliography{egbib}

\begin{thebibliography}{10}\itemsep=-1pt

\bibitem{barath2018graph}
Daniel Barath and Ji{\v{r}}{\'\i} Matas.
\newblock Graph-cut ransac.
\newblock In {\em CVPR}, 2018.

\bibitem{barath2019magsac}
Daniel Barath, Jiri Matas, and Jana Noskova.
\newblock Magsac: marginalizing sample consensus.
\newblock In {\em Proceedings of the IEEE Conference on Computer Vision and
  Pattern Recognition}, pages 10197--10205, 2019.

\bibitem{barath2020magsac++}
Daniel Barath, Jana Noskova, Maksym Ivashechkin, and Jiri Matas.
\newblock Magsac++, a fast, reliable and accurate robust estimator.
\newblock In {\em Proceedings of the IEEE/CVF Conference on Computer Vision and
  Pattern Recognition}, pages 1304--1312, 2020.

\bibitem{barroso2019key}
Axel Barroso-Laguna, Edgar Riba, Daniel Ponsa, and Krystian Mikolajczyk.
\newblock Key. net: Keypoint detection by handcrafted and learned cnn filters.
\newblock In {\em Proceedings of the IEEE International Conference on Computer
  Vision}, pages 5836--5844, 2019.

\bibitem{bian2017gms}
JiaWang Bian, Wen-Yan Lin, Yasuyuki Matsushita, Sai-Kit Yeung, Tan-Dat Nguyen,
  and Ming-Ming Cheng.
\newblock Gms: Grid-based motion statistics for fast, ultra-robust feature
  correspondence.
\newblock In {\em CVPR}, 2017.

\bibitem{black1993framework}
Michael~J Black and Padmanabhan Anandan.
\newblock A framework for the robust estimation of optical flow.
\newblock In {\em ICCV}, 1993.

\bibitem{black1996robust}
Michael~J Black and Paul Anandan.
\newblock The robust estimation of multiple motions: Parametric and
  piecewise-smooth flow fields.
\newblock {\em Computer vision and image understanding}, 63(1):75--104, 1996.

\bibitem{brachmann2017dsac}
Eric Brachmann, Alexander Krull, Sebastian Nowozin, Jamie Shotton, Frank
  Michel, Stefan Gumhold, and Carsten Rother.
\newblock Dsac-differentiable ransac for camera localization.
\newblock In {\em CVPR}, 2017.

\bibitem{brachmann2019ngransac}
Eric Brachmann and Carsten Rother.
\newblock {N}eural- {G}uided {RANSAC}: {L}earning where to sample model
  hypotheses.
\newblock In {\em ICCV}, 2019.

\bibitem{brown2007automatic}
Matthew Brown and David~G Lowe.
\newblock Automatic panoramic image stitching using invariant features.
\newblock {\em International journal of computer vision}, 74(1):59--73, 2007.

\bibitem{brox2010large}
Thomas Brox and Jitendra Malik.
\newblock Large displacement optical flow: descriptor matching in variational
  motion estimation.
\newblock {\em Transactions on pattern analysis and machine intelligence},
  33(3):500--513, 2010.

\bibitem{cavalli2020adalam}
Luca Cavalli, Viktor Larsson, Martin~Ralf Oswald, Torsten Sattler, and Marc
  Pollefeys.
\newblock Adalam: Revisiting handcrafted outlier detection.
\newblock {\em arXiv preprint arXiv:2006.04250}, 2020.

\bibitem{chen2019gla}
Zhi Chen, Fan Yang, and Wenbing Tao.
\newblock Gla-net: An attention network with guided loss for mismatch removal.
\newblock {\em ArXiv}, 2019.

\bibitem{Chum2005}
Ondrej Chum, Tomas Werner, and Jiri Matas.
\newblock Two-view geometry estimation unaffected by a dominant plane.
\newblock In {\em CVPR}, 2005.

\bibitem{darmon2020learning}
Fran{\c{c}}ois Darmon, Mathieu Aubry, and Pascal Monasse.
\newblock Learning to guide local feature matches.
\newblock {\em arXiv preprint arXiv:2010.10959}, 2020.

\bibitem{detone2018superpoint}
Daniel DeTone, Tomasz Malisiewicz, and Andrew Rabinovich.
\newblock Superpoint: Self-supervised interest point detection and description.
\newblock In {\em Proceedings of the IEEE Conference on Computer Vision and
  Pattern Recognition Workshops}, pages 224--236, 2018.

\bibitem{dusmanu2019d2}
Mihai Dusmanu, Ignacio Rocco, Tomas Pajdla, Marc Pollefeys, Josef Sivic,
  Akihiko Torii, and Torsten Sattler.
\newblock D2-net: A trainable cnn for joint detection and description of local
  features.
\newblock {\em arXiv preprint arXiv:1905.03561}, 2019.

\bibitem{fischler1981random}
Martin~A Fischler and Robert~C Bolles.
\newblock Random sample consensus: a paradigm for model fitting with
  applications to image analysis and automated cartography.
\newblock {\em Communications of the ACM}, 24(6):381--395, 1981.

\bibitem{hartley2003multiple}
Richard Hartley and Andrew Zisserman.
\newblock {\em Multiple view geometry in computer vision}.
\newblock Cambridge university press, 2003.

\bibitem{li2020dual}
Xinghui Li, Kai Han, Shuda Li, and Victor~Adrian Prisacariu.
\newblock Dual-resolution correspondence networks.
\newblock {\em arXiv preprint arXiv:2006.08844}, 2020.

\bibitem{lin2011smoothly}
Wen-Yan Lin, Siying Liu, Yasuyuki Matsushita, Tian-Tsong Ng, and Loong-Fah
  Cheong.
\newblock Smoothly varying affine stitching.
\newblock In {\em CVPR 2011}, pages 345--352. IEEE, 2011.

\bibitem{lin2017code}
Wen-Yan Lin, Fan Wang, Ming-Ming Cheng, Sai-Kit Yeung, Philip~HS Torr, Minh~N
  Do, and Jiangbo Lu.
\newblock Code: Coherence based decision boundaries for feature correspondence.
\newblock {\em Transactions on pattern analysis and machine intelligence},
  40(1):34--47, 2017.

\bibitem{lin2014bilateral}
Wen-Yan~Daniel Lin, Ming-Ming Cheng, Jiangbo Lu, Hongsheng Yang, Minh~N Do, and
  Philip Torr.
\newblock Bilateral functions for global motion modeling.
\newblock In {\em ECCV}, 2014.

\bibitem{liu2019gift}
Yuan Liu, Zehong Shen, Zhixuan Lin, Sida Peng, Hujun Bao, and Xiaowei Zhou.
\newblock Gift: Learning transformation-invariant dense visual descriptors via
  group cnns.
\newblock In {\em Advances in Neural Information Processing Systems}, pages
  6992--7003, 2019.

\bibitem{Lowe2004Distinctive}
David~G Lowe.
\newblock Distinctive image features from scale-invariant keypoints.
\newblock {\em International Journal of Computer Vision}, 60(2):91---110, 2004.

\bibitem{lucas1981iterative}
Bruce~D Lucas and Takeo Kanade.
\newblock An iterative image registration technique with an application to
  stereo vision.
\newblock {\em Proceedings DARPA image Understanding}, page 121430, 1981.

\bibitem{luo2019contextdesc}
Zixin Luo, Tianwei Shen, Lei Zhou, Jiahui Zhang, Yao Yao, Shiwei Li, Tian Fang,
  and Long Quan.
\newblock Contextdesc: Local descriptor augmentation with cross-modality
  context.
\newblock In {\em Proceedings of the IEEE Conference on Computer Vision and
  Pattern Recognition}, pages 2527--2536, 2019.

\bibitem{luo2018geodesc}
Zixin Luo, Tianwei Shen, Lei Zhou, Siyu Zhu, Runze Zhang, Yao Yao, Tian Fang,
  and Long Quan.
\newblock Geodesc: Learning local descriptors by integrating geometry
  constraints.
\newblock In {\em Proceedings of the European conference on computer vision
  (ECCV)}, pages 168--183, 2018.

\bibitem{ma2019locality}
Jiayi Ma, Ji Zhao, Junjun Jiang, Huabing Zhou, and Xiaojie Guo.
\newblock Locality preserving matching.
\newblock {\em International Journal of Computer Vision}, 127(5):512--531,
  2019.

\bibitem{mishchuk2017working}
Anastasiia Mishchuk, Dmytro Mishkin, Filip Radenovic, and Jiri Matas.
\newblock Working hard to know your neighbor's margins: Local descriptor
  learning loss.
\newblock In {\em Advances in Neural Information Processing Systems}, pages
  4826--4837, 2017.

\bibitem{moo2018learning}
Kwang Moo~Yi, Eduard Trulls, Yuki Ono, Vincent Lepetit, Mathieu Salzmann, and
  Pascal Fua.
\newblock Learning to find good correspondences.
\newblock In {\em CVPR}, 2018.

\bibitem{mur2015orb}
Raul Mur-Artal, Jose Maria~Martinez Montiel, and Juan~D Tardos.
\newblock Orb-slam: a versatile and accurate monocular slam system.
\newblock {\em Transactions on robotics}, 31(5):1147--1163, 2015.

\bibitem{myronenko2007non}
Andriy Myronenko, Xubo Song, and Miguel~A Carreira-Perpin{\'a}n.
\newblock Non-rigid point set registration: Coherent point drift.
\newblock In {\em NeurIPS}, 2007.

\bibitem{narayana2013coherent}
Manjunath Narayana, Allen Hanson, and Erik Learned-Miller.
\newblock Coherent motion segmentation in moving camera videos using optical
  flow orientations.
\newblock In {\em CVPR}, 2013.

\bibitem{ortega2018graph}
Antonio Ortega, Pascal Frossard, Jelena Kova{\v{c}}evi{\'c}, Jos{\'e}~MF Moura,
  and Pierre Vandergheynst.
\newblock Graph signal processing: Overview, challenges, and applications.
\newblock {\em Proceedings of the IEEE}, 106(5):808--828, 2018.

\bibitem{davis2017}
Jordi Pont-Tuset, Federico Perazzi, Sergi Caelles, Pablo Arbel\'aez, Alexander
  Sorkine-Hornung, and Luc {Van Gool}.
\newblock The 2017 davis challenge on video object segmentation.
\newblock {\em ArXiv}, 2017.

\bibitem{raguram2012usac}
Rahul Raguram, Ondrej Chum, Marc Pollefeys, Jiri Matas, and Jan-Michael Frahm.
\newblock Usac: a universal framework for random sample consensus.
\newblock {\em Transactions on pattern analysis and machine intelligence},
  35(8):2022--2038, 2012.

\bibitem{raguram2008comparative}
Rahul Raguram, Jan-Michael Frahm, and Marc Pollefeys.
\newblock A comparative analysis of ransac techniques leading to adaptive
  real-time random sample consensus.
\newblock In {\em ECCV}, 2008.

\bibitem{ranftl2018deep}
Ren{\'e} Ranftl and Vladlen Koltun.
\newblock Deep fundamental matrix estimation.
\newblock In {\em ECCV}, 2018.

\bibitem{revaud2019r2d2}
Jerome Revaud, Philippe Weinzaepfel, C{\'e}sar De~Souza, Noe Pion, Gabriela
  Csurka, Yohann Cabon, and Martin Humenberger.
\newblock R2d2: Repeatable and reliable detector and descriptor.
\newblock {\em arXiv preprint arXiv:1906.06195}, 2019.

\bibitem{rocco2020efficient}
Ignacio Rocco, Relja Arandjelovi{\'c}, and Josef Sivic.
\newblock Efficient neighbourhood consensus networks via submanifold sparse
  convolutions.
\newblock {\em arXiv preprint arXiv:2004.10566}, 2020.

\bibitem{rocco2018neighbourhood}
Ignacio Rocco, Mircea Cimpoi, Relja Arandjelovi{\'c}, Akihiko Torii, Tomas
  Pajdla, and Josef Sivic.
\newblock Neighbourhood consensus networks.
\newblock In {\em Advances in Neural Information Processing Systems}, pages
  1651--1662, 2018.

\bibitem{rousseeuw1984least}
Peter~J Rousseeuw.
\newblock Least median of squares regression.
\newblock {\em Journal of the American statistical association},
  79(388):871--880, 1984.

\bibitem{sarlin20superglue}
Paul-Edouard Sarlin, Daniel DeTone, Tomasz Malisiewicz, and Andrew Rabinovich.
\newblock {SuperGlue}: Learning feature matching with graph neural networks.
\newblock In {\em CVPR}, 2020.

\bibitem{sattler2018benchmarking}
Torsten Sattler, Will Maddern, Carl Toft, Akihiko Torii, Lars Hammarstrand,
  Erik Stenborg, Daniel Safari, Masatoshi Okutomi, Marc Pollefeys, Josef Sivic,
  et~al.
\newblock Benchmarking 6dof outdoor visual localization in changing conditions.
\newblock In {\em CVPR}, 2018.

\bibitem{song2020sekd}
Yafei Song, Ling Cai, Jia Li, Yonghong Tian, and Mingyang Li.
\newblock Sekd: Self-evolving keypoint detection and description.
\newblock {\em arXiv preprint arXiv:2006.05077}, 2020.

\bibitem{sun2020acne}
Weiwei Sun, Wei Jiang, Eduard Trulls, Andrea Tagliasacchi, and Kwang~Moo Yi.
\newblock Acne: Attentive context normalization for robust
  permutation-equivariant learning.
\newblock In {\em Proceedings of the IEEE/CVF Conference on Computer Vision and
  Pattern Recognition}, pages 11286--11295, 2020.

\bibitem{thomee2016yfcc100m}
Bart Thomee, David~A Shamma, Gerald Friedland, Benjamin Elizalde, Karl Ni,
  Douglas Poland, Damian Borth, and Li-Jia Li.
\newblock Yfcc100m: The new data in multimedia research.
\newblock {\em Communications of the ACM}, 59(2):64--73, 2016.

\bibitem{tian2020hynet}
Yurun Tian, Axel Barroso-Laguna, Tony Ng, Vassileios Balntas, and Krystian
  Mikolajczyk.
\newblock Hynet: Local descriptor with hybrid similarity measure and triplet
  loss.
\newblock {\em arXiv preprint arXiv:2006.10202}, 2020.

\bibitem{tian2019sosnet}
Yurun Tian, Xin Yu, Bin Fan, Fuchao Wu, Huub Heijnen, and Vassileios Balntas.
\newblock Sosnet: Second order similarity regularization for local descriptor
  learning.
\newblock In {\em Proceedings of the IEEE Conference on Computer Vision and
  Pattern Recognition}, pages 11016--11025, 2019.

\bibitem{tiwari2018dgsac}
Lokender Tiwari and Saket Anand.
\newblock Dgsac: Density guided sampling and consensus.
\newblock In {\em 2018 IEEE Winter Conference on Applications of Computer
  Vision (WACV)}, pages 974--982. IEEE, 2018.

\bibitem{torr2000mlesac}
Philip~HS Torr and Andrew Zisserman.
\newblock Mlesac: A new robust estimator with application to estimating image
  geometry.
\newblock {\em Computer vision and image understanding}, 78(1):138--156, 2000.

\bibitem{tyszkiewicz2020disk}
Micha{\l}~J Tyszkiewicz, Pascal Fua, and Eduard Trulls.
\newblock Disk: Learning local features with policy gradient.
\newblock {\em arXiv preprint arXiv:2006.13566}, 2020.

\bibitem{volz2011modeling}
Sebastian Volz, Andres Bruhn, Levi Valgaerts, and Henning Zimmer.
\newblock Modeling temporal coherence for optical flow.
\newblock In {\em CVPR}, 2011.

\bibitem{detrac}
Longyin Wen, Dawei Du, Zhaowei Cai, Zhen Lei, Ming{-}Ching Chang, Honggang Qi,
  Jongwoo Lim, Ming{-}Hsuan Yang, and Siwei Lyu.
\newblock {UA-DETRAC:} {A} new benchmark and protocol for multi-object
  detection and tracking.
\newblock {\em Computer Vision and Image Understanding}, 2020.

\bibitem{wiles2020d2d}
Olivia Wiles, Sebastien Ehrhardt, and Andrew Zisserman.
\newblock D2d: Learning to find good correspondences for image matching and
  manipulation.
\newblock {\em arXiv preprint arXiv:2007.08480}, 2020.

\bibitem{xiao2013sun3d}
Jianxiong Xiao, Andrew Owens, and Antonio Torralba.
\newblock Sun3d: A database of big spaces reconstructed using sfm and object
  labels.
\newblock In {\em ICCV}, 2013.

\bibitem{yi2016lift}
Kwang~Moo Yi, Eduard Trulls, Vincent Lepetit, and Pascal Fua.
\newblock Lift: Learned invariant feature transform.
\newblock In {\em European Conference on Computer Vision}, pages 467--483.
  Springer, 2016.

\bibitem{yuille1989mathematical}
Alan~L Yuille and Norberto~M Grzywacz.
\newblock A mathematical analysis of the motion coherence theory.
\newblock {\em International Journal of Computer Vision}, 3(2):155--175, 1989.

\bibitem{zhang2019learning}
Jiahui Zhang, Dawei Sun, Zixin Luo, Anbang Yao, Lei Zhou, Tianwei Shen, Yurong
  Chen, Long Quan, and Hongen Liao.
\newblock Learning two-view correspondences and geometry using order-aware
  network.
\newblock In {\em CVPR}, 2019.

\bibitem{zhao2019nm}
Chen Zhao, Zhiguo Cao, Chi Li, Xin Li, and Jiaqi Yang.
\newblock Nm-net: Mining reliable neighbors for robust feature correspondences.
\newblock In {\em CVPR}, 2019.

\end{thebibliography}
}

\end{document}